\documentclass[conference]{IEEEtran}
\IEEEoverridecommandlockouts
\usepackage{cite}
\usepackage{amsmath,amssymb,amsfonts}
\usepackage{algorithmic}
\usepackage{graphicx}
\usepackage{textcomp}
\usepackage{xcolor}
\usepackage{hyperref, enumitem}
\usepackage[caption=false,font=footnotesize]{subfig}
\usepackage{tikz}
\usetikzlibrary{positioning}
\usepackage{booktabs,threeparttable,multirow}

\def\BibTeX{{\rm B\kern-.05em{\sc i\kern-.025em b}\kern-.08em
    T\kern-.1667em\lower.7ex\hbox{E}\kern-.125emX}}

\begin{document}

\title{Traceback Translators Against Forgetting in Continual Fake Speech Detection \\
\thanks{This work was funded by the European Union under the Italian National Recovery and Resilience Plan (NRRP) of NextGenerationEU, with a partnership on “Telecommunications of the Future” (PE00000001 - program “RESTART”). 
We thank the partnership and collaboration with the S\~{a}o Paulo Research Foundation (Fapesp) Horus project, Grant \#2023/12865-8.
This manuscript reflects only the authors’ views and opinions, neither EU nor the European Commission nor FAPESP can be considered responsible for them. All of the authors have revised the document and confirm the findings.}
}

\author{\IEEEauthorblockN{Enrico Gottardis}
\IEEEauthorblockA{\textit{Department of Information Engineering} \\
\textit{University of Padova}\\
Padova, Italy \\
enrico.gottardis@studenti.unipd.it}
\and
\IEEEauthorblockN{Mattia Tamiazzo}
\IEEEauthorblockA{\textit{Department of Information Engineering} \\
\textit{University of Padova}\\
Padova, Italy \\
mattia.tamiazzo.1@phd.unipd.it}
\and
\IEEEauthorblockN{Simone Milani}
\IEEEauthorblockA{\textit{Department of Information Engineering} \\
\textit{University of Padova}\\
Padova, Italy \\
simone.milani@dei.unipd.it}
}

\maketitle

\begin{abstract}
Fake speech detectors are increasingly challenged by the development of new and more accurate generative models. 
To cope with this problem, continual learning techniques are nowadays widely considered feasible strategies for updating models to new datasets, but they also lead to decreased performance on previously seen samples (catastrophic forgetting). In this work, we propose a forgetting-resilient solution based on the adoption of domain translators within a frozen detector, which remaps the new feature spaces into the original ones by means of a traceback translator network. Experimental results show that this strategy enables the  achievement of high detection rates with respect to traditional retraining, while minimizing the computational effort and preserving the detection accuracy on previous data.
\end{abstract}


\section{Introduction and related works} \label{sec:introduction}

    In recent years, the widespread adoption and development of synthetic speech generators \cite{bhagtani2024purdue} have forced researchers and cybersecurity experts to create new deepfake detectors \cite{li2025survey}.

    As a matter of fact, existing detectors need to be updated through continual retraining in which old datasets are integrated with new samples, but unfortunately the access to the original data is not always guaranteed. 
    A possible alternative consists in fine-tuning pre-trained models on the sole novel dataset, but this leads to undesired learning drifts, such as catastrophic forgetting, where knowledge learned from old samples is diluted by new data.
    
    To address such issues, transfer learning, domain adaptation and continual learning techniques have been recently investigated in the literature \cite{wang2023lowrank,ba2023transferring,chen2025continual,zhang2024what,dong2024advancing,mulimani2024class,ma2021continual,negroni2025leveraging}.
    Methods based on transfer learning and domain adaptation aim at adapting pre-trained models and transferring knowledge across domains. These approaches include fine-tuning using Low-Rank Adaptation \cite{wang2023lowrank}, cross-lingual knowledge transfer \cite{ba2023transferring}, and unsupervised domain adaptation to handle distribution shifts \cite{chen2025continual}. 
    Strategies based on continual learning employ memory-based and regularization methods, such as self-adaptive learning to modify the gradient direction and retain previously learned knowledge \cite{zhang2024what}, and model distillation techniques \cite{dong2024advancing,mulimani2024class}. 
    The authors of \cite{ma2021continual} propose a continual learning method to incrementally train fake audio detectors on new spoofing attacks. The model includes a knowledge distillation loss and a similarity constraint that aligns genuine sample representations across different domains, improving detection performance. 
    
    To improve generalization across unseen data, a Mixture of Experts architecture is proposed in \cite{negroni2025leveraging}. Multiple specialized experts are trained on specific datasets, and their predictions are combined through a gating mechanism. By distributing specialized knowledge across multiple modules, the architecture is able to outperform single-model baselines in a cross-dataset scenario. 
    
    The study in \cite{salvi2025freeze} investigates the effectiveness of continual learning by comparing different retraining strategies. 
    The detector is divided into two logical sections (encoder and classification modules), and experiments evaluate the classification performance by applying a continual learning approach \cite{ma2021continual} to the entire model, or selectively to one section while freezing the other.
    Experimental results show that retraining the encoder while freezing the classifier offers a better balance between learning novel data and maintaining previously acquired knowledge.
    
    In this paper, we introduce a lightweight traceback domain translator into a frozen classifier in order to adapt it to new deepfake generators and sampling setups and mitigate catastrophic forgetting.
    It is possible to summarize the main contributions as follows:

    \begin{itemize}
    \item We propose a new continual learning approach for synthetic speech detection, where traditional fine-tuning is replaced by the introduction of a lightweight domain translator within a frozen classifier.
    \item The translator-based approach is compared to other domain-adaptation solutions; the number of retrained parameters is minimized while maximizing accuracies on both new and old datasets.
    \item The proposed solution was tested in a multilingual setup (including both English and Chinese sentences), showing that the proposed approach can also be used for a cross-lingual adaptation.
    \item Performance was further improved by means of generative data augmentation based on diffusion models.
    \end{itemize}

\section{Continual learning and catastrophic forgetting in synthetic speech detection models}
    In order to tackle the problem of forgetting in continual learning for synthetic audio classification, we tested different adaptation methods measuring their impacts on a simple but effective classifier.

    \begin{figure}[t]
        \centering
        \includegraphics[width=\linewidth]{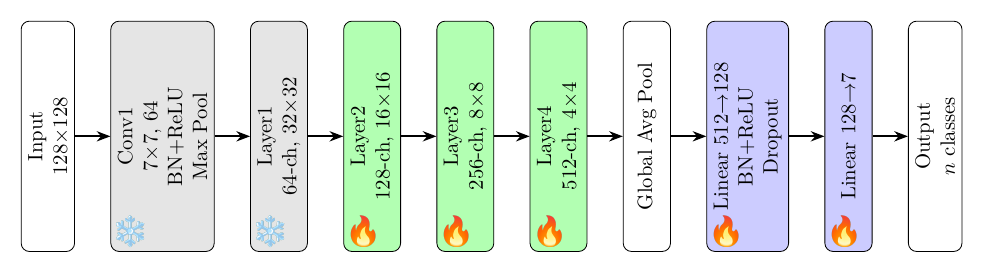}
        \caption{Architecture of the customized ResNet18 used in the work. }
        \label{ResNet18_original_architecture}
    \end{figure}

    \subsection{Dataset pre-processing}
    The adopted architecture processes the input signal and generates sets of  MFCC spectrograms including $128$-frequency coefficients and spanning a time window of $42$ frames around salient instants (leading to $128 \times 42$ matrices). Salient instants are selected at points where the signal evolves from a silenced interval to a voiced one since their relevance has already been highlighted in some previous works \cite{mari2022sound, mari2025all}.
    
    Although multiple spectrograms can be extracted from a single file, for some classes the number of available samples was limited, leading to a high class imbalance between real and synthetic audio samples. To address this issue, we resorted to a dataset generative process based on a diffusion model implemented in \cite{hf_unet}. 
    Specifically, we employed a 2D U-Net architecture \cite{ronneberger2015unet} as the core denoising component of the diffusion process, enabling the generation of synthetic spectrogram samples while preserving the spatial structure and dimensions of the input representations.
    To further compensate for the class imbalance, classic data augmentation was used, and an imbalanced sampler was adopted. 
    
    \subsection{Audio deepfake detection}
        The classification model is based on a customized ResNet18 \cite{he2016resnet} (see \figurename~\ref{ResNet18_original_architecture}), where the first convolutional layer is frozen and the classifier head consists of two fully-connected layers with normalization and dropout.

        The model was trained using a weighted custom Cross-Entropy loss with label smoothing, to reduce over-confidence:
        \begin{equation}
            \begin{aligned}
                &\mathcal{L}_{\mathrm{CE}} =
                - \frac{1}{N} \sum_{i=1}^{N} \sum_{c=0}^{K-1} w_c \, \tilde{y}_{i,c} \, \log \hat{p}_{i,c}, 
                \\
                & \hspace{1.1cm} \tilde{y}_{i,c} =
                \begin{cases} 
                    1-\epsilon, & c = y_i \\[1mm]
                    \epsilon/K, & c \neq y_i
                \end{cases}
            \end{aligned}
        \end{equation}
        where $\mathbf{w} = [6,1,1,1,1,1,1] \in \mathbb{R}^K$ is the vector of weights, with $w_c$ the $c$-th element of $\mathbf{w}$, $\epsilon = 0.08$, $K=7$ represents the number of classes, and $N$ is the batch size. 
        From experimental results, we noticed a tendency of the model to misclassify real samples as if they were synthetic from class $6$. For this reason, we added an extra penalty defined as:
        \begin{equation}
            \mathcal{L}_{\mathrm{penalty}} = 
            \lambda \sum_{i=1}^{N} \mathbf{1}\big(y_i = 0 \ \wedge \ \arg\max_c \hat{p}_{i,c} = 6\big),
        \end{equation}
        with $\lambda = 5.0$, leading to a combined classifier loss:
        \begin{equation}
            \mathcal{L}_{\mathrm{classifier}} = \mathcal{L}_{\mathrm{CE}} + \mathcal{L}_{\mathrm{penalty}}.\label{classifier_loss}
        \end{equation}
        For validation, we employed a standard Cross-Entropy loss.

    \subsection{Continual Learning Strategies} \label{subsec:cl_strategies}
        After obtaining a solid backbone model to perform detection on audio deepfakes, we analyzed different approaches to perform continual learning on other datasets. 
        
        \subsubsection{Full retraining}
            The first approach was to retrain the whole network on the new dataset. This common method is largely implemented by other researchers \cite{salvi2025freeze}, leading to impressive performance. However, the main drawback is the \textit{catastrophic forgetting} phenomenon: the model learns the new representation of data, forgetting the previous task.
        \subsubsection{Domain adaptation}
            In order to reduce the catastrophic forgetting issue, we opted to retrain only part of the whole architecture. In particular, we considered adopting a domain adaptation approach by selectively retraining only the Batch Normalization (BN) layers, while keeping the rest of the model frozen (sometimes called Continual Normalization \cite{Pham2022ContinualNR}). 
            This method has two main advantages: the retraining of the model requires only the new dataset, on which the network will be fine-tuned, and the training will be faster, as the only learnable parameters are the BN layers. 
            
        \subsubsection{Domain Translation}
            This idea consists of shifting the distributions of real and fake samples in the target domain to make them similar to those in the source domain. 

            Although the proposed translator shares similarities with adapter modules explored in~\cite{houslby2019parameter,kessler2022adapter,eeckt2023using}, its role and characteristics are different. The translator is applied after the embedding layer and is trained to perform class-conditional alignment of real or fake feature distributions across domains. By freezing the backbone and adapting only this lightweight module, the approach limits catastrophic forgetting. To the best of our knowledge, the use of simple adapter-like architectures has not been previously explored for distribution alignment in audio deepfake detection.

\begin{figure}[t]
                \centering
                \includegraphics[width=0.85\linewidth]{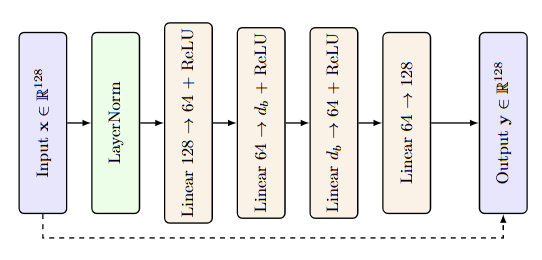}
                \vspace{-2ex}
                \caption{Block diagram for the translator model.}
                \label{Translator_architecture}
            \end{figure}

            We considered the features created from the model after the embedding, which provides vectors in a $128$-dimensional latent space. Before providing these vectors to the fully connected part of the model, we processed them with the architecture described in Fig.~\ref{Translator_architecture}, where a fully-connected model with a bottleneck $d_b=32$ is followed by a residual connection $\mathbf{y} = \mathbf{x} + f(\mathrm{LN}(\mathbf{x}))$, shown in the picture with a dashed line.
            To promote the translation process, the training loss consists of three elements: the classifier loss described in~\eqref{classifier_loss}, the CORrelation ALignment (CORAL) loss~\cite{sun2016deep} and Prototype Consistency (PC)~\cite{zhang2025prototypical} loss. 
            While the CORAL loss aims at minimizing the mean and variance between real and fake samples from the source and the target datasets, PC minimizes the intra-class distance.
            Given source features $z_s \in \mathbb{R}^{N_B \times 128}$ with labels $y_s \in \{0,\dots,6\}$ and target features $z_t \in \mathbb{R}^{N_B \times 128}$ with labels $y_t \in \{0,1\}$, with $N_B$ being the batch size, we define their batch-wise means  $\mu_s^{\text{real}}$, $\mu_t^{\text{real}}$, $\mu_s^{\text{fake}}$, $\mu_t^{\text{fake}}$ as class prototypes and the corresponding covariance matrices as $\Sigma_s$ and $\Sigma_t$. 
            Let us define the norm $\lVert \mu_s - \mu_t \rVert_2^2=\Big(\| \mu_s^{\text{real}} - \mu_t^{\text{real}} \|_2^2 + \| \mu_s^{\text{fake}} - \mu_t^{\text{fake}} \|_2^2\Big)$.
            
            The \emph{prototype consistency loss} is defined as:
            \begin{equation}
                \mathcal{L}_{\mathrm{PC}} = 
                \frac{1}{2} \lVert \mu_s - \mu_t \rVert_2^2.
            \end{equation}
            Denoting $\lVert \cdot \rVert_F^2$ as the squared matrix Frobenius norm, it is then possible to compute the \emph{CORAL loss} as:
            \begin{equation}
                \mathcal{L}_{\mathrm{CORAL}} = 
                \underbrace{\lVert \mu_s - \mu_t \rVert_2^2}_{\text{mean alignment}} 
                + \underbrace{\lVert \Sigma_s - \Sigma_t \rVert_F^2}_{\text{covariance alignment}}.
            \end{equation}
            The final training loss is defined as:
            \begin{equation}
            \mathcal{L}_{\mathrm{total}} = 
            \mathcal{L}_{\mathrm{classifier}} +
            \lambda_{CORAL} \,\mathcal{L}_{\mathrm{CORAL}} + \lambda_{PC} \,\mathcal{L}_{\mathrm{PC}}\label{eq:final_loss}
            \end{equation}
            where $\lambda_{CORAL}$ and $\lambda_{PC}$ are two regularization terms, experimentally set to $0.05$ and $0.01$ respectively.
            
\section{Experimental setup}
    The experimental setting for our continual learning approach was defined based on several state-of-the-art corpora of samples.
    
    \begin{itemize}[leftmargin=0pt, label={}]
        \item \textbf{ASVspoof 2019 (ASV19)} \cite{todisco2019asvspoof}.
        The dataset was introduced to assess countermeasures against synthetic and voice-converted speech in English and is organized into three subsets: train, development, and evaluation. It is one of the most widely used benchmark datasets for audio deepfake detection.
        \item \textbf{FakeOrReal (FoR)} \cite{reimano2019for}.
        The dataset contains English bona fide and synthetic speech generated by various Text-To-Speech (TTS) systems, including both open-source and commercial tools. The real speech content is sourced from publicly available speech datasets.
        \item \textbf{In-The-Wild (ITW)} \cite{muller2022itw}. 
        This corpus aims to assess speech detectors in a real-world scenario and includes real and synthetic audio samples collected from English-speaking celebrities and politicians. Covering diverse recording conditions and generative methods, it is particularly useful for evaluating model robustness.
        \item \textbf{ADD 2022 (ADD22)} \cite{yi2022add}. 
        The dataset includes genuine and synthetic Mandarin Chinese audio samples. It is designed to evaluate deepfake detection systems across various conditions, such as low-quality and partially fake audio tracks.
    \end{itemize}

    The dataset used to train the baseline model is the training subset of ASVspoof 2019. 
    Although training was performed on individual spectrograms, inference results were aggregated at the audio-excerpt level using majority voting over multiple spectrograms extracted from the same sample.

\section{Results}
    This section presents a preliminary backbone selection study, followed by the evaluation of the proposed continual learning strategies, and an ablation analysis.
    
    \subsection{Backbone Architecture Selection}
    Before evaluating the proposed continual learning strategies, we performed a preliminary study to identify a suitable backbone model for the detection task.
    Several architectures were considered, evaluating them on the ASVspoof 2019 development set.
    
    Table~\ref{tab:model_comparison} reports the comparison among the investigated architectures. 
    Among the evaluated models, the customized ResNet18 provided the best performance even compared to more complex architectures, like EfficientNet~\cite{efficientnet2019tan}, MobileNet~\cite{mobilenet2019howard}, and ConvNeXT-Tiny~\cite{convnet2022liu}.
    The Audio Spectrogram Transformer (AST)~\cite{ast2021gong} achieved comparable accuracy; however, it requires substantially higher computational resources, with approximately 86 million trainable parameters compared to the 11 million parameters of ResNet18, as well as longer training times. 

    \begin{table}[t]
        \centering
        \caption{Model comparison for classification task on ASVspoof 2019.}
        \label{tab:model_comparison}
        \begin{tabular}{lccc}
            \toprule
            Architecture & Dev. accuracy & Dev. loss  & Tr. epochs \\
            \midrule
            \textbf{Custom ResNet18} & \textbf{0.994} & \textbf{0.112}  & \textbf{240} \\
            AST~\cite{ast2021gong} & 0.994 & 0.193  & 324 \\
            ConvNeXT-Tiny~\cite{convnet2022liu} & 0.904 & 0.329  & 296 \\
            ResNet18 - multiclass~\cite{he2016resnet} & 0.883 & 0.269  & 148 \\
            ResNet18 - binary~\cite{he2016resnet} & 0.860 & 0.398  & 110 \\
            EfficientNet B0~\cite{efficientnet2019tan} & 0.823 & 0.461 & 60 \\
            EfficientNet B1~\cite{efficientnet2019tan} & 0.776 & 0.536  & 30 \\
            MobileNet V3 Large~\cite{mobilenet2019howard} & 0.753 & 0.588 & 100 \\
            \bottomrule
        \end{tabular}
        \vskip -2ex
    \end{table} 

    \begin{table*}[t]
        \centering
        \caption{Experimental results of the proposed methods. Results in brackets are referred to the performance on the ASVspoof 2019 evaluation set after specialization (retraining). The method is compared with the LCNN-based model in \cite{salvi2025freeze}.}
        \label{tab:results}
        \begin{tabular}{lcccccccc}
            \toprule
            \multirow{2}{*}{Method} & \multicolumn{2}{c}{ASV19} & \multicolumn{2}{c}{FOR} & \multicolumn{2}{c}{ITW} & \multicolumn{2}{c}{ADD22} \\
            \cmidrule(lr){2-3} \cmidrule(lr){4-5} \cmidrule(lr){6-7} \cmidrule(lr){8-9}
            & AUC (\%) & EER (\%) & AUC (\%) & EER (\%) & AUC (\%) & EER (\%) & AUC (\%) & EER (\%) \\
            \midrule
            Baseline model & 95.0 & 9.74 & 49.8 & 50.2 & 49.6 & 50.3 & 30.8 & 62.9 \\
            Full retraining & -- & -- & 99.7 \textcolor{blue}{(61.2)} & 3.27 \textcolor{blue}{(43.2)} & 100 \textcolor{blue}{(59.6)} & 0.28 \textcolor{blue}{(43.8)} & 100 \textcolor{blue}{(22.1)} & 0.78 \textcolor{blue}{(71.8)} \\
            Domain adaptation & -- & -- & 97.4 \textcolor{blue}{(63.1)} & 8.26 \textcolor{blue}{(40.7)} & 96.3 \textcolor{blue}{(77.6)} & 9.15 \textcolor{blue}{(28.8)} & 99.3 \textcolor{blue}{(46.0)} & 4.02 \textcolor{blue}{(52.2)} \\
            Domain translation & -- & -- & 96.0 \textcolor{blue}{(95.0)} & 11.37 \textcolor{blue}{(9.74)} & 95.4 \textcolor{blue}{(95.0)} & 10.64 \textcolor{blue}{(9.74)} & 98.1 \textcolor{blue}{(95.0)} & 6.96 \textcolor{blue}{(9.74)} \\
            \midrule
            CL ALL \cite{salvi2025freeze} & 95.7 & 11.1 & 99.0 \textcolor{blue}{(94.0)} & 4.20 \textcolor{blue}{(13.6)} & 99.8 \textcolor{blue}{(92.1)} & 1.60 \textcolor{blue}{(15.4)} & -- & -- \\
            CL $\mathcal{M}_c$ \cite{salvi2025freeze} & -- & -- & 98.8 \textcolor{blue}{(93.2)} & 5.00 \textcolor{blue}{(14.0)} & 99.5 \textcolor{blue}{(93.1)} & 2.80 \textcolor{blue}{(14.6)} & -- & -- \\
            CL $\mathcal{M}_e$ \cite{salvi2025freeze} & -- & -- & 98.9 \textcolor{blue}{(93.2)} & 4.90 \textcolor{blue}{(13.4)} & 99.0 \textcolor{blue}{(91.8)} & 4.10 \textcolor{blue}{(15.1)} & -- & -- \\
            \bottomrule
        \end{tabular}
    \end{table*}
    
    \begin{figure*}[t]
        \centering
        \vskip -1ex
        \subfloat[ROC plots for the full retraining of the model.]{%
               \includegraphics[width=.24\linewidth]{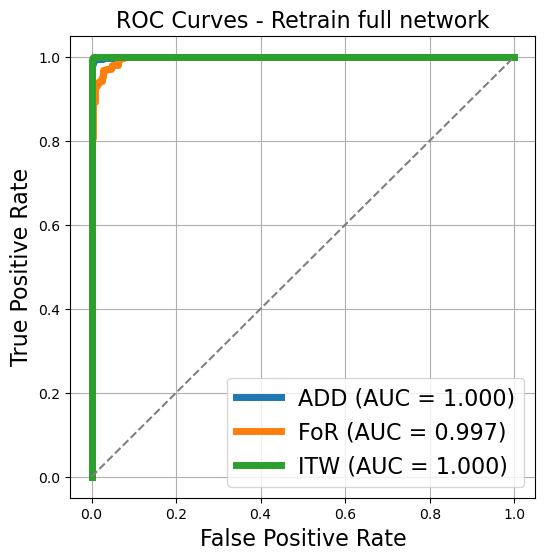}%
                \label{fig:ROCs_retrain_all}} 
        \hfil
        \subfloat[ROC plots for the domain adaptation approach.]{%
               \includegraphics[width=.24\linewidth]{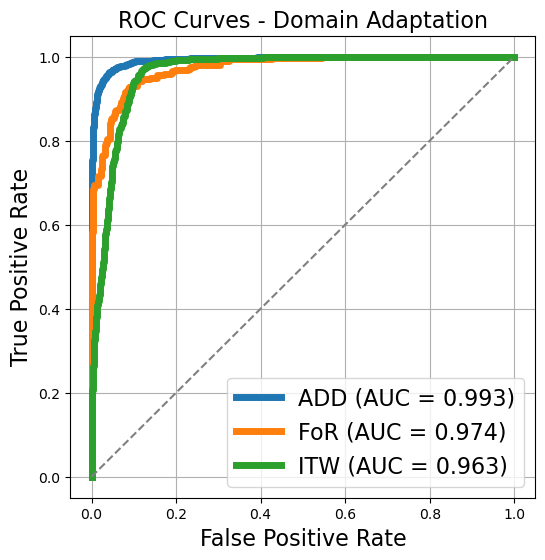}%
                \label{fig:ROCs_BN_retraining}} 
        \hfil
        \subfloat[ROC plots for the domain translation approach.]{%
               \includegraphics[width=.24\linewidth]{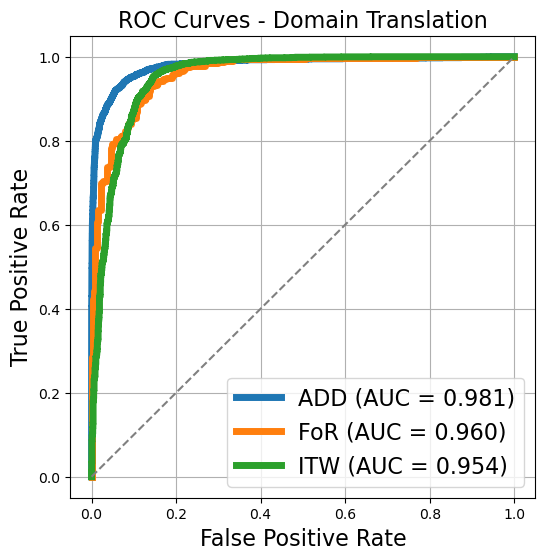}%
                \label{fig:ROCs_domain_translation}} 
        \caption{Comparison of the ROC plots obtained testing the proposed methods.}
        \label{fig:ROCs_combined}
    \end{figure*}

        \begin{table}[t]
            \centering
            \caption{Number of parameters of our methods compared with the LCNN model employed in \cite{salvi2025freeze}.}
            \label{tab:comparison_numbers_parameters}
            \begin{tabular}{lccc}
                \toprule
                Method & \# Trained parameters  \\
                \midrule
                Full retraining    & 11095K \\
                Domain adaptation  & 10K  \\
                Domain translation & 21K   \\ 
                \midrule
                CL ALL \cite{salvi2025freeze} &  5556K \\
                CL $\mathcal{M}_c$ \cite{salvi2025freeze} & 521K  \\
                CL $\mathcal{M}_e$ \cite{salvi2025freeze} & 5035K \\
                \bottomrule
            \end{tabular}
            \vskip -2ex
        \end{table}

    \subsection{Main Experimental Results}
    The results obtained on the evaluation datasets are summarized in Table \ref{tab:results}. The baseline model achieves performance comparable to that reported in \cite{salvi2025freeze}, evaluated using the Receiver Operating Characteristic (ROC) curve, and the Equal Error Rate (EER). Performance is also quantified using the Area Under the Curve (AUC). 
    The results of the three continual learning strategies introduced in \ref{subsec:cl_strategies} are illustrated in Fig.\ref{fig:ROCs_combined}.
    It can be observed that full retraining performs very well on the target dataset (average $1.44\%$ EER on new dataset), but results in a dramatic loss of accuracy on the source dataset ASVspoof 2019 ($-52\%$ AUC and $+52.9\%$ EER on average). The largest loss is reasonably observed on ADD since the language change from English to Chinese. Performing a partial update (domain adaptation) proved to be more robust, as the average AUC decrease is $2.3\%$ on the target domain, but the loss on the source domain is $38\%$ AUC and EER increases by $30.8\%$ on average. It is also worth noting that only a small number of parameters must be retrained (see Table~\ref{tab:comparison_numbers_parameters}), making it very efficient.
    The domain translation strategy represents a good trade-off between classification performances and computational complexity
    since the average AUC loss on target domains is $3.4\%$, but we preserve the accuracy on the source domain ($95.4\%$ AUC, $9.74\%$ EER) while updating only a limited number of parameters.
    
    The experimental results also show that the proposed approaches can be applied to cross-lingual adaptation, since ADD dataset includes Chinese language speech samples. The traceback translation model proves to be capable of handling datasets in different languages while keeping the same classification accuracy.

    \subsection{Ablation studies}
        To determine the impact of diffusion-based data augmentation on model performance, we trained the baseline model on only the ASVspoof 2019 training set, and then we included both original and generated samples. Experimental results show consistent performance gains with data augmentation: classification accuracy increased from $81.9\%$ to $83.7\%$, while the AUC improved from $91.5\%$ to $95.0\%$.
        
        Moreover, we enabled/disabled the different components in~\eqref{eq:final_loss} and evaluated their impact on the final accuracy. With only $\mathcal{L}_{\mathrm{PC}}$ (Prototype-Consistency), EER for FoR/ITW/ADD increases to $10.75\%$, $11.87\%$, $7.34\%$ respectively, while using  $\mathcal{L}_{\mathrm{CORAL}}$ only (CORAL loss), the respective EER is $11.84\%$, $9.05\%$, $7.72\%$, showing that their combination is, on average, more effective on all the datasets.

\section{Conclusions}
In this paper, we investigated the effectiveness of multiple continual learning techniques for audio deepfake detection to mitigate the problem of catastrophic forgetting. 

Our experiments showed that all the approaches achieve good detection performance, but the full retraining and domain adaptation methods are not able to retain previously learned knowledge. 
We proposed a novel approach based on domain translation that achieves high detection rates without requiring the retraining of the backbone architecture and, at the same time, preserves the detection accuracy on the original dataset. 

Future work will focus on extending the evaluation to a larger number of datasets to test scalability and robustness against different types of domain shifts. Additionally, exploring different model architectures could further improve performance in cross-domain and cross-lingual audio deepfake detection.

\section*{Acknowledgment}
The authors would like to thank Davide Salvi, Viola Negroni, Luca Bondi, Paolo Bestagini, and Stefano Tubaro of Politecnico di Milano for their support in the experimental comparison with \cite{salvi2025freeze}.

\bibliographystyle{IEEEtran}
\bibliography{refs}

\end{document}